# Performance Comparisons of PSO based Clustering

[1]Suresh Chandra Satapathy, [2]Gunanidhi Pradhan, [3]Sabyasachi Pattnaik, [4]JVR Murthy, [5]PVGD Prasad Reddy
[1]Anil Neerukonda Institute of Technology and Sciences, Sangivalas,Vishakapatnam Dist
[2]Bhubanananda Orissa School of Engineering, Cuttack
[3]FM University, Balasore
[4]JNTU College of Engineering, Kakinda,
[5]College of Engineering, AU, Vishakapatnam,

**Abstract**

**In this paper we have investigated the performance of PSO Particle Swarm Optimization based clustering on few real world data sets and one artificial data set. The performances are measured by two metric namely quantization error and inter-cluster distance. The K means clustering algorithm is first implemented for all data sets, the results of which form the basis of comparison of PSO based approaches. We have explored different variants of PSO such as *gbest, lbest ring, lbest vonneumann* and *Hybrid PSO* for comparison purposes. The results reveal that PSO based clustering algorithms perform better compared to K means in all data sets.**

## 1. INTRODUCTION

Data clustering is the process of grouping together similar multi-dimensional data vectors into a number of clusters or bins. Clustering algorithms have been applied to a wide range of problems, including exploratory data analysis, data mining [1], image segmentation [2] and mathematical programming [3,4] Clustering techniques have been used successfully to address the scalability problem of machine learning and data mining algorithms. Clustering algorithms can be grouped into two main classes of algorithms, namely supervised and unsupervised. With supervised clustering, the learning algorithm has an external teacher that indicates the target class to which a data vector should belong. For unsupervised clustering, a teacher does not exist, and data vectors are grouped based on distance from one another. This paper focuses on unsupervised clustering.

Many unsupervised clustering algorithms have been developed one such algorithm is K-means which is simple, straightforward and is based on the firm foundation of analysis of variances. The main drawback of the K-means algorithm is that the result is sensitive to the selection of the initial cluster centroids and may converge to the local optima. This is solved by PSO as it performs globalized search for solutions.

So this paper explores the applicability of PSO and its variants to cluster data vectors. In the process of doing so, the objective of the paper is:

- to show that the standard PSO algorithm can be used to cluster arbitrary data, and
- to compare the performance of PSO and its variants with standard K-means algorithm.

The rest of the paper is organized as follows: Section 2 presents an overview of K-means algorithm. The basic PSO and its variants are discussed in section 3. Function optimization using PSO models are given in section 4. How Clustering is done with PSO is discussed in section 5. Experimental results are summarized in section 6 and Conclusion and further work is emphasized in section 7.

## 2. K-Means Clustering

One of the most important components of a clustering algorithm is the measure of similarity used to determine how close two patterns are to one another. K-means clustering group data vectors into a predefined number of clusters, based on Euclidean distance as similarity measure. Data vectors within a cluster have small Euclidean distances from one another, and are associated with one centroid vector, which represents the "midpoint" of that cluster. The centroid vector is the mean of the data vectors that belong to the corresponding cluster.

For the purpose of this paper, following symbols are defined:

- $N_d$ denotes the input dimension, i.e. the number of parameters of each data vector
- $N_o$ denotes the number of data vectors to be clustered
- $N_c$ denotes the number of cluster centroids (as provided by the user), i.e. the number of clusters to be formed

- $z_p$ denotes the p[th] data vector
- $m_j$ denotes the centroid vector of cluster j
- $n_j$ is the number of data vectors in cluster j
- $C_j$, is the subset of data vectors that form cluster j.

Using the above notation, the standard K-means algorithm is summarized as

1. Randomly initialize the $N_c$ cluster centroid vectors
2. Repeat

(a) For each data vector, assign the vector to the class with the closest centroid vector, where the distance to the centroid is determined using

$$d(z_p, m_j) = \sqrt{\sum_{k=1}^{N_d} (z_{pk} - m_{jk})^2} \quad \text{---- (1)}$$

Where ***k*** subscripts the dimension

(b) Recalculate the cluster centroid vectors, using

$$m_j = \frac{1}{n_j} \sum_{\forall z_p \in C_j} z_p \quad \text{----- (2)}$$

until a stopping criterion is satisfied,

The K-means clustering process can be stopped when any one of the following criteria are satisfied: when the maximum number of iterations has been exceeded, when there is little change in the centroid vectors over a number of iterations or when there are no cluster membership changes. For the purposes of this study, the algorithm is stopped when a user-specified number of iterations have been exceeded.

## 3. Particle Swarm Optimization and its variants

Particle swarm optimization (PSO) is a population-based stochastic search process, modeled after the social behavior of a bird flock [5,6]. The algorithm maintains a population of particles, where each particle represents a potential solution to an optimization problem. In the context of PSO, a swarm refers to a number of potential solutions to the optimization problem, where each potential solution is referred to as a particle. The aim of the PSO is to find the particle position that results in the best evaluation of a given fitness (objective) function.

Each particle represents a position in $N_d$ dimensional space, and is: “flown” through this multi-dimensional search space, adjusting its position toward both

- the particle's best position found thus far. and
- the best position in the neighborhood of that particle.

Each particle ***i*** maintains the following information:

- $x_i$ : The ***current*** *position* of the particle;
- $v_i$ : The ***current velocity.*** of the particle;
- $y_i$ : The ***personal best position*** of the panicle.

Using the above notation. a particle's position is adjusted according to

$$v_{i,k}(t+1) = wv_{i,k}(t) + c_1 r_{1,k}(t)(y_{i,k}(t) - x_{i,k}(t)) + c_2 r_{2,k}(t)\left(\hat{y}_{i\,k}(t) - x_{i,k}(t)\right) \quad \text{---------- (3)}$$

$$x_i(t+1) = x_i(t) + v_i(t+1) \quad \text{------- (4)}$$

$r_{1,j}(t), r_{2,j}(t) \sim U(0,1)$ and k=1,….., $N_d$

Where $w$ is the inertia weight, $c_1, c_2$ are the acceleration constants and r is the random number generated for avoiding and biasing effect to social and cognitive components.

The velocity is thus calculated based on three contributions: (1) a fraction of the previous velocity, (2) the cognitive component which is a function of the distance of the particle from its personal best position, and (3) the social component which is a function of the distance of the particle from the best particle found thus far (i.e. the best of the personal bests)

The personal best position of particle $i$ is calculated as

$$y_i(t+1) = \begin{cases} y_i(t) & \text{if } f(x_i(t+1)) \geq f(y_i(t)) \\ x_i(t+1) & \text{if } f(x_i(t+1)) < f(y_i(t)) \end{cases} \quad (5)$$

Two basic approaches to PSO exist based on the interpretation of the neighborhood of particles. Equation (3) reflects the *gbest* version of PSO where, for each particle, the neighborhood is simply the entire swarm. The social component then causes particles to be drawn towards the best particle in the swam. In the *lbest* PSO model, the swam is divided into overlapping neighborhoods, and the best particle of each neighborhood is determined. For the *lbest* PSO model, the social component of equation (3) changes to.

$$c_2 r_{2,k}(t)\left(\hat{y}_{j,k}(t) - x_{i,k}(t)\right) \quad (6)$$

Where $\hat{y}_j$ is the best particle in the neighborhood of the $i^{th}$ particle.

The PSO is usually executed with repeated application of equations (3) and (4) until a specified number of iterations have been exceeded. Alternatively, the algorithm can be terminated when the velocity updates are close to zero over a number of iterations. *lbest-ring* is one of the variant of PSO in which the pbest is determined with respect to the neighboring adjacent particles as shown in figure 1.

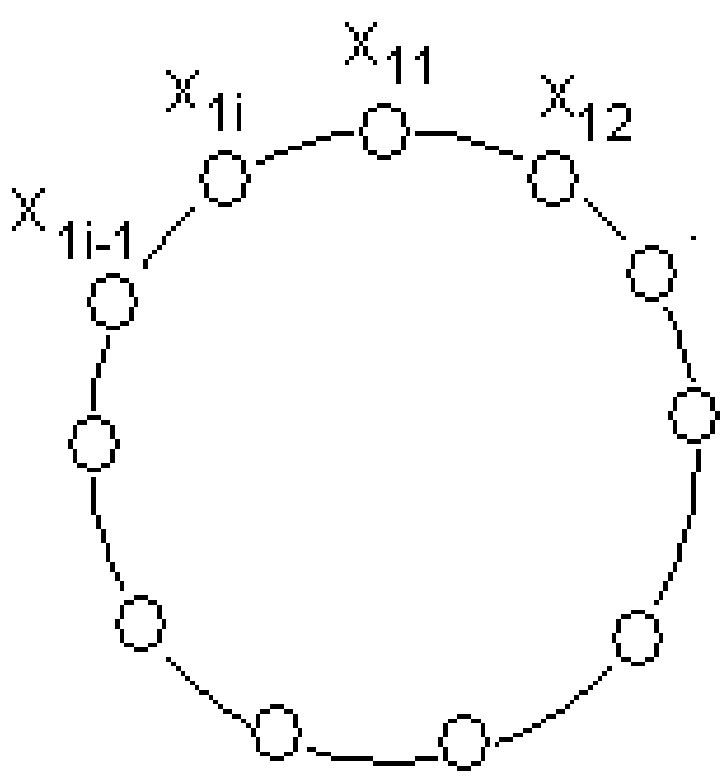

Figure 1 - Ring architecture

In Von-Neumann architecture the particles are considered to be in two dimensional matrix. pbest of the particle is determined with respect to four neighboring adjacent particles as shown in figure 2.

| | | | | | |
|---|---|---|---|---|---|
| $X_{11}$ | $X_{12}$ ↑ | * | * | $X_{1i-1}$ | $X_{1i}$ |
| $X_{21}$ ← | $X_{22}$ → | * | * | $X_{2i-1}$ | $X_{2i}$ |
| $X_{31}$ | ↓ $X_{32}$ | * | * | $X_{3i-1}$ | $X_{3i}$ |
| * | * | * | * | * | * |
| * | * | * | * | * | * |
| $X_{j1}$ | $X_{j2}$ | * | * | $X_{ji-1}$ | $X_{ji}$ |

Figure 2 - Von-Neumann architecture

**4 PSO a tool for Function Optimization**

PSO can be applied to number of real world problems like optimization which has been expanding in all directions at an astonishing rate. So the optimization of several complex functions is done with PSO. We have applied the different variations of PSO namely *lbest (ring and von-Neumann architectures)* [7,8] and *gbest* for optimizing some standard Benchmark functions given in the Table I [7], with range of search and maximum velocities in Table II, and corresponding results are given in table III.

Table I: Benchmarks for simulations

| function | Mathematical representation |
|---|---|
| Sphere function | $f_1(x) = \sum_{i=1}^{n} x_i^2$ |
| RosenBrock function | $f_2(x) = \sum_{i=1}^{n-1} [100(x_i^2 - x_{i+1})^2 + (x_i - 1)^2]$ |
| Rastrigrin function | $f_3(x) = 10n + \sum_{i=1}^{n} [x_i^2 - 10\cos(2\Pi x_i)]$ |

Table II: Range of search and Maximum Velocity

| Function | Range of search | Maximum Velocity |
|---|---|---|
| $f_1$ | $(-100,100)^n$ | 100 |
| $f_2$ | $(-100,100)^n$ | 100 |
| $f_3$ | $(-10,10)^n$ | 10 |

Table-III: Results

| Function | Type of solving | Dimension | Iterations | Best fitness | Mean | Standard deviation |
|---|---|---|---|---|---|---|
| f1 | gbest | 10 | 3000 | 1.000000e+005 | 9.999201e+004 | 5.620078e+001 |
| | | 20 | 4000 | 2.000000e+005 | 1.973950e+005 | 4.119254e+003 |
| | | 30 | 5000 | 3.000000e+005 | 2.863525e+005 | 1.275881e+004 |
| | lbest-ring | 10 | 3000 | 1.000000e+005 | 1.000000e+005 | 0 |
| | | 20 | 4000 | 2.000000e+005 | 2.000000e+005 | 6.067948e-011 |
| | | 30 | 5000 | 3.000000e+005 | 2.999999e+005 | 8.273009e-009 |
| | lbest-VonNeumann | 10 | 3000 | 1.000000e+005 | 1.000000e+005 | 0 |
| | | 20 | 4000 | 2.000000e+005 | 1.999998e+005 | 1.261541e+000 |
| | | 30 | 5000 | 3.000000e+005 | 2.999999e+005 | 6.746770e-008 |
| f2 | gbest | 10 | 3000 | 9.060897e+010 | 8.960774e+010 | 4.532346e+008 |
| | | 20 | 4000 | 1.906303e+011 | 1.873680e+011 | 4.039168e+009 |
| | | 30 | 5000 | 2.909030e+011 | 2.755630e+011 | 1.199402e+010 |
| | lbest-ring | 10 | 3000 | 9.140909e+010 | 9.084980e+010 | 3.239145e+008 |
| | | 20 | 4000 | 1.922191e+011 | 1.904858e+011 | 7.115941e+008 |
| | | 30 | 5000 | 2.927112e+011 | 2.857646e+011 | 5.024448e+009 |
| | lbest-VonNeumann | 10 | 3000 | 9.140909e+010 | 9.047086e+010 | 4.351285e+008 |
| | | 20 | 4000 | 1.921883e+011 | 1.901236e+011 | 1.385913e+009 |
| | | 30 | 5000 | 2.918292e+011 | 2.848440e+011 | 6.836449e+009 |
| f3 | gbest | 10 | 3000 | 1.107131e+003 | 1.106769e+003 | 2.559094e+000 |
| | | 20 | 4000 | 2.214210e+003 | 2.175897e+003 | 3.953340e+001 |
| | | 30 | 5000 | 3.313335e+003 | 3.173358e+003 | 1.009656e+002 |
| | lbest-ring | 10 | 3000 | 1.107131e+003 | 1.105852e+003 | 3.251735e+000 |
| | | 20 | 4000 | 2.214262e+003 | 2.160460e+003 | 3.290293e+001 |
| | | 30 | 5000 | 3.263586e+003 | 3.147861e+003 | 5.840186e+001 |
| | lbest-VonNeumann | 10 | 3000 | 1.107131e+003 | 1.106766e+003 | 2.558340e+000 |
| | | 20 | 4000 | 2.214263e+003 | 2.184117e+003 | 3.052733e+001 |
| | | 30 | 5000 | 3.321395e+003 | 3.205129e+003 | 7.052384e+001 |

From the above results it can be seen that the PSO is a very good candidate for solving optimization problems. So the data clustering problem is a sort of optimization problem where in the objective is to find a similar data objects into a specific group. In our work the PSO is used for investigating this objective.

## 5. PSO Clustering

In the context of clustering, a single particle represents the $N_c$ cluster centroid vectors. That is, each particle $x_i$ is constructed as follows:

$$\mathbf{x}_i = (\mathbf{m}_{i1}, \cdots, \mathbf{m}_{ij}, \cdots, \mathbf{m}_{iN_c}) \qquad (7)$$

where $\mathbf{m}_{ij}$ refers to the j-th cluster centroid vector of the i-th particle in cluster $C_{ij}$. Therefore, a swarm represents a number of candidate clusters for the current data vectors. The fitness of particles is easily measured as the quantization error,

$$J_e = \frac{\sum_{j=1}^{N_c} [\sum_{\forall \mathbf{z}_p \in C_{ij}} d(\mathbf{z}_p, \mathbf{m}_j)/|C_{ij}|]}{N_c} \qquad (8)$$

Where $d$ is defined in equation (1), and $|C_{ij}|$ is the number of data vectors belonging to cluster $C_{ij}$, i.e. the frequency of that cluster.

This section presents a standard PSO for clustering data into a given number of clusters.

### 5.1 PSO Cluster Algorithm

Using the standard *gbest* PSO, data vectors can be clustered as follows:

1. Initialize each particle to contain $N_c$, randomly selected cluster centroids.
2. For $t = 1$ to $t_{max}$ do
   (a) For each particle *i* do

(b) For each data vector $z_p$

i) Calculate the Euclidean distance $d(z_p, m_{i,j})$ to all cluster centroids $C_{ij}$

ii) Assign $z_p$ to cluster $C_{ij}$ such that

$$d(\mathbf{z}_p, \mathbf{m}_{ij}) = \min_{\forall c=1,\cdots,N_c}\{d(\mathbf{z}_p, \mathbf{m}_{ic})\}$$

iii) Calculate the fitness using equation (8)

(c) Update the global best and local best positions
(d) Update the cluster centroids using equations (3) and (4)

Where *t,,,* is the maximum number of iterations.

The population-based search of the PSO algorithm reduces the effect that initial conditions have, as opposed to the K-means algorithm; the search starts from multiple positions in parallel. Section 6 shows that the PSO algorithm performs better than the K-means algorithm in terms of quantization error.

$$\begin{array}{ccccccc}
X_{11} & X_{12} & X_{13} & ------- & X_{1i} & ------ & X_{1m} \\
X_{21} & X_{22} & X_{13} & ------- & X_{2i} & ------ & X_{2m} \\
X_{31} & X_{32} & X_{33} & ------- & X_{3i} & ------ & X_{3m} \\
| & | & | & ------- & | & ------ & | \\
| & | & | & ------- & | & ------ & | \\
| & | & | & ------- & | & ------ & | \\
| & | & | & ------- & | & ------ & | \\
| & | & | & ------- & | & ------ & | \\
X_{n1} & X_{n2} & X_{n3} & ------- & X_{ni} & ------ & X_{nm}
\end{array}$$

## 6. Data Set and Experimental Results

This section compares the results of the K-means and PSO algorithms on five clustering problems. The main purpose is to compare the quality of the respective clustering, where quality is measured according to the following two criteria:

- the quantization error as defined in equation (8);
- the inter-cluster distances, i.e. the distance between the centroids of the clusters, where the objective is to maximize the distance between clusters.

For all the results reported, averages over 30 simulations are given. All algorithms are run for 1000 function evaluations, and the PSO algorithms used 10 particles. The Hybrid PSO takes the seed from result of K-means clustering. This seed is considered as one particle in swarm of particles in PSO. For PSO, w is varying as per the paper [9]. The initial weight is fixed at 0.9 and the final weight at 0.4. The acceleration coefficients c1 and c2 are fixed at 1.042 to ensure good convergence [10].

The clustering problems used for the purpose of this paper are:

- Iris plants database: This is a well-understood database with **4** inputs, **3** classes and 150 data vectors.
- Wine: This is a classification problem with "well behaved" class structures. There are 13 inputs, ***3*** classes and **178** data vectors.
- Hayes Roth which has **132** data vectors with 3 classes and 5 inputs.
- Diabetes data set has 768 data vectors having 2 classes and 8 inputs.
- Artificial: This problem follows the following classification rule;

$$class1 = 1 \quad \text{if} \quad \begin{array}{l}(z_1 \geq 0.7) or ((z_1 \leq 0.3) \\ and (z_2 \geq -0.2 - z_1))\end{array}$$

$$class2 = 0 \quad \text{Otherwise}$$

A total of 400 data vectors are randomly created between (-1,1).

Table IV summarizes the results obtained from the five clustering algorithms for the problems cited above. The values reported are averages over *30* simulations, with standard deviations to indicate the range of values to which the algorithms converge. First, consider the fitness of solutions, i.e. the quantization error, for all data sets PSO based clustering is better than K-means. However, *lbest*_VonNuumann provides better fitness values in terms of quantization error and inter_cluster distance for all data sets except for Wine. For Wine and Hayes Roth , Hybrid PSO gives good result. The lbest_Vonneuumann gives worst quantization error but comparatively good inter_cluster distance measure for these data sets. The standard deviations (std) found to be very close, thereby indicating the convergence of algorithms to better results.

Table IV: Results of clustering

| Data Sets | Algorithm | Quantization error,std | Inter – cluster distance,std |
|---|---|---|---|
| Iris | K means | 0.733, 0.21075 | 8.3238, 1.6821 |
| | PSO_gbest | 0.027209, 0.017964 | 19.41, 2.4693 |
| | lbest_ring | 0.026615, 0.014664 | 21.079, 3.8171 |
| | lbest_vonneumann | 0.012477, 0.019458 | 20.278, 0.02204 |
| | Hybrid PSO | 0.68743, 0.019081 | 18.598, 0.65266 |
| Hayes Roth | K means | 11.961, 1.573 | 8.9353, 1.2419 |
| | PSO_gbest | 0.77086, 0.0408 | 324.25, 5.7895 |
| | lbest_ring | 3.99,3.3429 | 313.1, 3.4562 |
| | lbest_vonneumann | 3.8265, 0.98856 | 350.73, 23.272 |
| | Hybrid PSO | 0.57914, 1.9488 | 323.51, 61.738 |
| Wine | K means | 116.29, 0.83715 | 2019.2,0.234 |
| | PSO_gbest | 10.765, 3.7278 | 3272.8, 292.89 |
| | lbest_ring | 33.622, 7.6328 | 2859.7, 339.91 |
| | lbest_vonneumann | 11.709, 1.6749 | 3450.8, 222.42 |
| | Hybrid PSO | 3.9663, 4.3043 | 3596.7, 483.11 |
| Diabetes | K means | 78.984, 7.6654 | 20.92, 3.332 |
| | PSO_gbest | 69.222, 2.4839 | 30.12,2.719 |
| | lbest_ring | 36.98, 2.397 | 36.108, 2.475 |
| | lbest_vonneumann | 33.205, 2.8501 | 38.1074,2.4714 |
| | Hybrid PSO | 48.545,3.097 | 32.958,3.471 |
| Artificial | K means | 0.64152, 0.011867 | 1.3772, 0.02580 |
| | PSO_gbest | 0.54338, 0.0057227 | 1.2678, 0.72643 |
| | lbest_ring | 0.56021, 0.004647 | 1.482, 0.13267 |
| | lbest_vonneumann | 0.5317,0.00121 | 1.662,0.11 |
| | Hybrid PSO | 0.55086,0.00056684 | 0.9086, 0.16526 |

## 7. CONCLUSION

This paper investigates the application of the PSO to cluster data vectors. Five algorithms were tested, namely a standard K-means, *gbest* PSO, lbest_ring, lbest_vonneummann and Hybrid PSO. The PSO approaches are compared against K-means clustering, which showed that the PSO approaches have better convergence to lower quantization errors, and in general, larger inter-cluster distances. Future studies will involve more elaborate tests on higher dimensional problems-and large number of patterns. The PSO clustering algorithms will also be extended to dynamically determine the optimal number of clusters.